\documentclass[11pt]{article}
\usepackage{acl2016}
\usepackage{times}
\usepackage{latexsym}
\usepackage{url}
\usepackage{color}
\usepackage{multirow}
\usepackage{amssymb}
\usepackage[dvips]{graphicx}
\usepackage{epstopdf}
\usepackage{algorithmicx}
\usepackage[noend]{algpseudocode}
\usepackage{algorithm}
\usepackage{mathtools}
\usepackage{amsthm}

\usepackage{amsmath}

\algnewcommand{\LeftComment}[1]{\Statex \(\triangleright\) #1}

\newcommand{\changed}[1]{\textcolor{black}{#1}}
\newcommand{\concept}[1]{\textbf{#1}}
\newcommand{\changedtwo}[1]{\textcolor{black}{#1}}
\newcommand{\myfunctionname}[1]{{\operatorname{\mathit{#1}}}} %why not both? even nicer
\newcommand{\quotedtext}[1]{{\emph{\textquoteleft {#1}\textquoteright}}}

\aclfinalcopy % Uncomment this line for the final submission
\newtheorem{mydef}{Definition}
\newtheorem{myexmpl}{Example}

\title{Universal, Unsupervised \changedtwo{(Rule-Based)}, Uncovered Sentiment Analysis\Thanks{NOTICE: this is the authors version of a work that was accepted for publication in Knowledge-Based Systems. Changes resulting from the publishing process, such as peer review, editing, corrections, structural formatting, and other quality control mechanisms may not be reflected in this document. Changes may have been made to this work since it was submitted for publication. A definitive version has been published in Knowledge-Based systems (http://dx.doi.org/10.1016/j.knosys.2016.11.014). \textcopyright 2016. This manuscript version is made available under the CC-BY-NC-ND 4.0 license http://creativecommons.org/licenses/by-nc-nd/4.0/ }}

\author{David Vilares, Carlos G\'{o}mez-Rodr\'{\i}guez and Miguel A. Alonso\\
	Grupo LyS, Departamento de Computaci\'{o}n, Universidade da Coru\~{n}a \\
         Campus de A Coru\~{n}a s/n, 15071, A Coru\~{n}a, Spain \\
  {\tt \{david.vilares,carlos.gomez,miguel.alonso\}@udc.es}}

\date{}

\begin{document}



% \title{Universal, Unsupervised \changedtwo{(Rule-Based)}, Uncovered Sentiment Analysis}


% % \ead{carlos.gomez@udc.es}
% 
% \author{Miguel A. Alonso}\corref{cor3}\fnref{label_miguel}}
% \ead{miguel.alonso@udc.es}

\maketitle

\begin{abstract}

We present a novel unsupervised approach for multilingual sentiment analysis driven by compositional syntax-based rules.
On the one hand, we exploit some of the main advantages of unsupervised algorithms: (1) the interpretability of their output, in contrast with most supervised models, which behave as a black box and (2) their robustness across different corpora and domains.
On the other hand, by introducing the concept of compositional operations and exploiting syntactic information in the form of universal dependencies, we tackle one of their main drawbacks: their rigidity on data that are structured differently depending on the language concerned.
Experiments show an improvement both over existing unsupervised methods, and over state-of-the-art supervised models 
when evaluating outside their corpus of origin. \changed{Experiments also show how the same compositional operations can be shared across languages.}
The system is available at \changed{\url{http://www.grupolys.org/software/UUUSA/}}. 

\end{abstract}


\section{Introduction}

Sentiment Analysis ({\sc sa}) is a subfield of natural language processing ({\sc nlp}) that deals with the automatic comprehension of the opinions shared by users in different media \cite{Pang2008,cambria2016affective}. One of the main challenges addressed by {\sc sa} focuses on emulating the semantic composition process carried out by humans when understanding the sentiment of an opinion (i.e., if it is favorable, unfavorable or neutral).
In the sentence \emph{\textquoteleft He is not very handsome, but he has something that I really like\textquoteright}, humans have the ability to infer that
the word \emph{\textquoteleft very\textquoteright} emphasizes \emph{\textquoteleft handsome\textquoteright}, \emph{\textquoteleft not\textquoteright} affects the whole expression \emph{\textquoteleft very handsome\textquoteright},
and \emph{\textquoteleft but\textquoteright} decreases the relevance of \emph{\textquoteleft He is not very handsome\textquoteright} and increases the one of \emph{\textquoteleft he has something that I really like\textquoteright}. Based on this, a human could justify a positive overall sentiment on that sentence.

Our main contribution is the introduction of the first universal and unsupervised (knowledge-based) model for compositional sentiment analysis ({\sc sa}) driven by
syntax-based rules. We introduce a formalism for compositional operations, allowing the creation of arbitrarily complex rules to tackle relevant phenomena for {\sc sa}, for any language and syntactic dependency annotation.  We implement and evaluate a set of practical universal operations defined using part-of-speech (PoS) tags and dependency types under the universal guidelines of \newcite{PetDasMcD2011a} and \newcite{McdNivQuirGolDasGanHalPetZhaTacBedCasLee2013}: universal annotation criteria that can be used to represent the morphology and syntax of any language in a uniform way. The model outperforms existing unsupervised approaches as well as state-of-the-art
compositional supervised models \cite{SocPerWuChuManNgPot2013a} on domain-transfer settings, and shows that the operations can be shared across languages, as they are defined using universal guidelines. %\changed{, universal annotation criteria that can be used to represent the morphology and syntax of any language in a uniform way, and have been used to annotate a number of corpora.}

% The code and the system are freely available for the research community at \url{anonoymous-url}.
% Additionally, an interactive demo can be found in: \url{anonymous-url}.

% However, in supervised \emph{sentiment analysis} ({\sc sa}) and \emph{polarity classification}, hierarchical semantic composition is often ignored,
% % Traditional machine learning models first, and more recently deep neural networks, address the task as a statistical word co-occurrence problem, 
% although there are exceptions \cite{SocPerWuChuManNgPot2013a}. 

% In spite of being powerful and accurate, supervised approaches also present drawbacks. Firstly, they behave as a black box. Secondly, they do not perform so well on domain transfer applications \cite{Aue2005,Pang2008}. Thirdly, they are often trained on a single language. Finally, feature and hyper-parameter engineering can be time and resource costly options.
% When these limitations need to be addressed, unsupervised (rule-based) approaches are useful. However, only a few unsupervised approaches
% are able to tackle multiple languages, and those that can do not apply semantic composition either \cite{Thelwall:2012}.
% 

% The remainder of the paper is organized as follows. \S \ref{section-related-work} describes related work, focusing on syntactic and unsupervised
% approaches. \S \ref{section-unsupervised-analysis} describes our unsupervised approach. \S \ref{section-experiments} shows and discusses the experimental results. Finally, \S \ref{section-conclusions} draws our conclusions and future ideas.

The remainder of this article is structured as follows. \S \ref{section-related-work} reviews related work. \S \ref{section-unsupervised-analysis} introduces the formalism for compositional operations, which is used in \S \ref{section-real-compositional-operations} to define a set of universal rules that can process relevant linguistic phenomena for {\sc sa} in any language. \S \ref{section-experiments} presents experimental results of our approach on different corpora and languages. Finally, \S \ref{section-conclusions} concludes and discusses directions for future work.  

\section{Related work}\label{section-related-work}

In this section we describe previous work relevant to the topics covered in this article: the issue of multilinguality in {\sc sa}, semantic composition through machine learning models and semantic composition on knowledge-based systems.

\subsection{Multilingual SA}

Monolingual sentiment analysis systems have been created for languages belonging to a variety of language families, such as Afro-Asiatic~\cite{AldAzm2015a}, Indo-European~\cite{VilAloGom2015a,VilAloGom2014a,GhoJac2011a,SchCon2013a,NerAliCapCuaBy2012a,HabPtaSte2014a,MedShaWha2013a,MedShaWha2013a}, Japonic~\cite{AraKamAizSuz2014a}, Sino-Tibetan~\cite{VinCha2012a,ZhaZenLiWanZuo2009a} and Tai-Kadai~\cite{InrSin2010a}, among others.

The performance of a given approach for sentiment analysis varies from language to language. In the case of supervised systems, the size of the training set is a relevant factor~\cite{CheZhu2013a,DemPec2013a}, but performance is also affected by linguistic particularities~\cite{BoiMou2009a,Wan2009a} and the availability of language processing tools~\cite{KliCim2014a} and resources~\cite{SevMosUryPlaFil2016a}. With respect to the latter point, sentiment lexicons are scarce for languages other than English, and therefore a great deal of effort has been dedicated to building lexical resources for sentiment analysis~\cite{KimJunNamLeeLee2009a,HogHeeFraKayJon2014a,CruTroPonOrt2014b,VolWilYar2013a,GaoWeiLiLiuZho2013a,CheSki2014a}. A common approach for obtaining a lexicon for a new language consists in translating pre-existent English lexicons~\cite{BroTofTab2009a}, but it was found that even if the translation is correct, two parallel words do not always share the same semantic orientation across languages 
due to 
differences in common usage~\cite{GhoJac2011a}.

Another approach for building a monolingual {\sc sa} system for a new language is based on the use of machine translation ({\sc mt}) in order to translate the text into English automatically, to then apply a polarity classifier for English, yielding as a result a kind of cross-language sentiment analysis system~\cite{BalTur2012a,Wan2009a,PerMarUreMar2013a,MarMarMolPer2014a}. It was found that text with more sentiment is harder to translate than text with less sentiment~\cite{CheZhu2014a} and that translation errors produce an increase in the sparseness of features, a fact that degrades performance~\cite{BalTur2012b,BalTur2014a}. To deal with this issue, several methods have been proposed to reduce translation errors, such as applying both directions of translation simultaneously~\cite{HajIbrSel2014a} or enriching the {\sc mt} system with sentiment patterns~\cite{HirTetHid2004a}. In the case of supervised systems, self-training and co-training techniques have also been explored to improve 
performance~\cite{GuiXuXuYuaYaoShoQiuWanWonChe2013a,GuiXuLuXuXuLiuWan2014a}. 

Few multilingual systems for {\sc sa} tasks have been described in the literature. Banea et al.~\cite{BanMihWie2010a,BanMihWie2014a} describe a system for detecting subjectivity (i.e., determining if a text contains subjective or objective information) in English and Romanian texts, finding that 90\% of word senses maintained their subjectivity content across both languages. Xiao and Guo~\cite{XiaGuo2012a} confirm on the same dataset that boosting on several languages improves performance for subjectivity classification with respect to monolingual methods. 

Regarding the few multilingual polarity classification systems described in the literature, they are based  on a supervised setting. In this respect, Yan et al.~\cite{YaHeSheTan2014a} describe a supervised multilingual system for {\sc sa} working on previously tokenized Chinese and English texts. \newcite{VilAloGomWASSA2015} present a multilingual {\sc sa} system trained on a multilingual dataset that is able to outperform monolingual systems on some monolingual datasets and that can work successfully on code-switching texts, i.e., texts that contain terms written in two or more different languages~\cite{VilAloGom2016a}. Some approaches rely on {\sc mt} to deal with multi-linguality. Balahur et al.~\cite{BalTurStePerJacKucZavElG2014a} build a supervised multilingual {\sc sa} system by translating the English SemEval 2013 Twitter dataset~\cite{ChoGueTonLav2013a} into other languages by means of {\sc mt}, which improves on the results of monolingual systems due to the fact that, when multiple languages are 
used 
to 
build 
the classifier, the features that are relevant are automatically selected. They also point out that the performance of the monolingual Spanish {\sc sa} system trained on Spanish machine translated data can be improved by adding native Spanish data for training from the Spanish TASS 2013 Twitter dataset~\cite{VilGarLarGon2014a}. In contrast, Balahur and Perea-Ortega~\cite{BalPer2015a} inform that performance decreases when machine-translated English data is used to enlarge the TASS 2013 training corpus for Spanish sentiment analysis. 

Other approaches advocate the use of language-independent indicators of sentiment, such as emoticons~\cite{DavGha2011a}, for building language-independent {\sc sa} systems, although the accuracy of a system built following this approach is worse than the combined accuracy of monolingual systems~\cite{NarHulAlb2012a}. The use of other language-independent indicators, such as character and punctuation repetitions, results in low recall~\cite{CuiZhaLiuMa2011a}.

\subsection{Composition in machine learning SA systems}

A na\"{\i}ve approach to emulating the comprehension of the meaning of multiword phrases for {\sc sa}
consists in using \emph{n}-grams of words, with $n>1$ \cite{Pang-ML}. The approach is limited by the curse of dimensionality, although crawling data from the target domain can help to reduce that problem \cite{KirShuMoh2014a}.
\newcite{JosPen2009a} went one step forward and proposed generalized dependency triplets as features for subjectivity detection, capturing non-local relations.
\newcite{socher2012semantic} modeled a recursive neural network that learns compositional
vector representations for phrases and sentences of arbitrary syntactic type and length.
\newcite{SocPerWuChuManNgPot2013a} presented an improved recursive deep model for {\sc sa} over dependency
%trees. To train it, they created a sentiment treebank
%tagged using Amazon Mechanical Turk, pushing the state of the art up to 85.4\% on the \newcite{PanLee2005} dataset.
trees, and trained it on a sentiment treebank tagged using Amazon Mechanical Turk, pushing the state of the art up to 85.4\% on the Pang and Lee 2005 dataset \cite{PanLee2005}.
\newcite{KalGreBlu2014a} showed how convolutional neural networks ({\sc cnn}) can be used for semantic modeling of sentences. The model implicitly captures local and non-local relations without the need of a parse tree. It can be adapted for any language, as long as enough data is available.
\newcite{SevMos2015} showed the effectiveness of a {\sc cnn} in a SemEval sentiment analysis shared task \cite{RosNakKirMohManSto2015}, although crawling tens of millions of messages was first required to achieve state-of-the-art results. \changed{With a different purpose, \newcite{poria2016aspect} presented a deep learning approach for aspect extraction in opinion mining, classifying the terms of a sentence as aspect or non-aspect. The system is then enriched with linguistic patterns specifically defined for aspect-detection tasks, which helps improve the overall performance and shows the utility of combining supervised and rule-based approaches.}

In spite of being powerful and accurate, supervised approaches like these also present drawbacks. Firstly, they behave as a black box. Secondly, they do not perform so well on domain transfer applications \cite{Aue2005,Pang2008}. Finally, feature and hyper-parameter engineering can be time and resource costly options.

\subsection{Composition in knowledge-based SA systems}

When the said limitations of machine learning models need to be addressed, unsupervised approaches are useful. 
In this line, \newcite{Turney2002a} proposed an unsupervised learning algorithm to calculate the semantic orientation ({\sc so}) of a word.
\newcite{Lexicon-BasedMethods} presented a lexical rule-based approach to handle relevant linguistic phenomena such as intensification, negation, \emph{\textquoteleft but\textquoteright} clauses and \emph{irrealis}.
\newcite{Thelwall:2012} released SentiStrength, a multilingual unsupervised system for micro-text {\sc sa} that handles negation and intensification, among other web linguistic phenomena. It is limited to snippet-based and word-matching rules, since no {\sc nlp} phases such as part-of-speech tagging or parsing are applied. Regarding syntax-based approaches, the few described in the literature are language-dependent. 
\newcite{JiaYuMen2009a} define a set of syntax-based rules for handling negation in English. \newcite{VilAloGom2015a} propose a syntactic {\sc sa} method, but limited to Spanish reviews and Ancora trees \cite{Ancora}.
\newcite{CamOlsDhee2014} release SenticNet v3, a resource for performing sentiment analysis in English texts at the semantic level
rather than at the syntactic level, by combining existing resources such as ConceptNet \cite{LiuSing2004} and AffectiveSpace \cite{CamHusHavEck2009}.
By exploiting artificial intelligence ({\sc ai}), semantic web technologies and dimensionality reduction techniques it computes
the polarity of multiword common-sense concepts (e.g. \texttt{buy Christmas present}).
With a different goal, \newcite{LiuGaoLiuZha2016a} automatically select syntactical rules for an unsupervised aspect extraction approach, showing the utility of rule-based systems on opinion mining tasks.

In brief, most unsupervised approaches are language-dependent, and those that can manage multilinguality, such as SentiStrength, cannot apply semantic composition.

\section{Unsupervised Compositional SA}\label{section-unsupervised-analysis}

In contrast with previous work, we propose a formalism for compositional operations, allowing the creation of arbitrarily complex rules to tackle relevant phenomena for {\sc sa}, for any language and syntactic dependency annotation.

\subsection{Operations for compositional SA}\label{section-operations-compositional-sa}
% \subsection{Part-of-speech tagging}

Let $w$=$w_{1},...,w_{n}$ be a sentence, where each word occurrence $w_{i} \in W$.

\begin{mydef}
A {\bf tagged sentence} is a list of tuples $(w_{i},t_{i})$ where each $w_{i}$ is assigned a part-of-speech tag, $t_{i}$, indicating its grammatical category (e.g. noun, verb or adjective).
\end{mydef}

% \begin{myexmpl}
% Given our running example: \textquoteleft He is not very handsome, but he has something that I really like\textquoteright, a PoS tagged sentence would be:
%  [(He, pronoun), (is, verb), (not, adverb), (very, adverb), (handsome, adjective), (but, conjuction), (he, pronoun), (has, verb), (something, adverb), (that, preposition) (I, pronoun), (really, adverb), (like, verb)].\footnote{Tags assigned following the guidelines proposed by \newcite{PetDasMcD2011a}}
% \end{myexmpl}

% Our approach sets an alternative to state-of-art supervised approaches that obtain high results on the corpus where they were trained, but that often have problems on domain-transfer applications.
% \subsection{Dependency trees}

%Let $w$=$w_{1},...,w_{n}$ be an input string with $w_{i} \in W$ and where each $w_{i}$ is assigned to a part-of-speech (PoS) tag $t_{i} \in T$. 
% Let $w$=$w_{1},...,w_{n}$ be a sentence, where each word occurrence $w_{i} \in W$ is assigned a PoS tag $t_{i} \in T$.

\begin{mydef}

A {\bf dependency tree} for w is an edge-labeled directed tree $T=(V,E)$ where $V = \{0,1,2,\ldots,n\}$ is the set of nodes and $E = V \times D \times V$ is the set of labeled arcs. Each arc, of the form $(i,d,j)$, corresponds to a syntactic {\bf dependency} between the words $w_i$ and $w_j$; where $i$ is the index of the {\bf head} word, $j$ is the index of the {\bf child} word and $d$ is the {\bf dependency type} representing the kind of syntactic relation between them. Following standard practice, we use node $0$ as a dummy root node that acts as the head of the syntactic root(s) of the sentence.

% A {\bf dependency tree} for w is an edge-labeled directed tree $T=(V,E)$ where $V = \{0,1,2,\ldots,n\}$ is the set of nodes and $E = V \times D \times V$ is the set of labeled arcs. Each arc, of the form $(i,d,j)$, corresponds to a syntactic {\bf dependency} between the words $w_i$ and $w_j$; where $i$ is the index of the {\bf head} or {\bf parent} word, $j$ is the index of the {\bf dependent} or {\bf child} word and $d$ is the {\bf dependency type} representing the kind of syntactic relation between them. Following standard practice, we use node $0$ as a dummy root node that acts as the head of the syntactic root(s) of the sentence.

\end{mydef}

\begin{myexmpl}
Figure \ref{figure-running-example-raw} shows a valid dependency tree for our running example.
 \begin{figure*}[hbpt]
   \centering
  \includegraphics[width=13.5cm,clip]{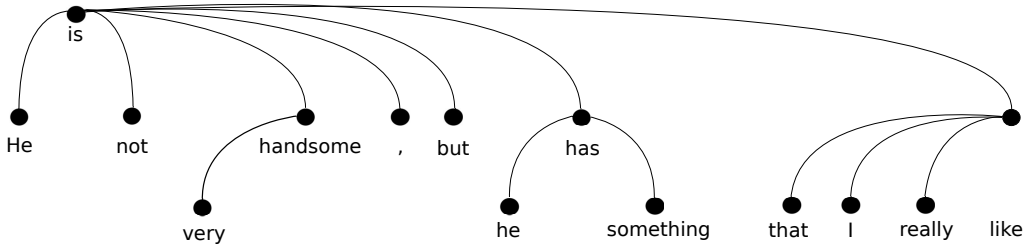}
  \caption{Example of a valid dependency tree for our introductory sentence: \emph{\textquoteleft He is not very handsome, but he has something that I really like\textquoteright}, following the \newcite{McdNivQuirGolDasGanHalPetZhaTacBedCasLee2013} guidelines. For simplicity, we omit the dummy root in the figures.}
  \label{figure-running-example-raw}
\end{figure*}

\end{myexmpl}

We will write $i \xrightarrow{d} j$ as shorthand for $(i,d,j) \in E$ and
% When a given tree $T=(V,E)$ is understood by the context, we will also write $i \xrightarrow{d} j$ to mean that there is an arc $i \xrightarrow{d} j \in E$; and 
we will omit the dependency types when they are not relevant.
%\noindent We denote as a {\bf dependency triplet} to $(i,d,j)\in E$,  where $i$ is the {\bf head} or {\bf parent} term, $j$ is the {\bf dependent} or {\bf child} term and $d$
%is the {\bf dependency type} from the $i$ to $j$ that represents the existing syntactic relation between them. 
Given a dependency tree $T=(V,E)$, and a node $i \in V$, we define a set of functions to obtain the context of node $i$:
\begin{itemize}
\itemsep-0.1em
\item $\myfunctionname{ancestor}_T(i,{\delta})=\{ k \in V :$ there is a path of length $\delta \text{ from } k \text{ to } i \text{ in } T \}$, i.e., the singleton set containing the ${\delta}$th ancestor of $i$ (or the empty set if there is no such node),

% $\myfunctionname{ancestor}_T(i,{\delta}) = \{ k \in V : \text{ there is a path of length } \delta \text{ from } k \text{ to } i \text{ in } T \}$. 
% For example, if $i \rightarrow j, \, j \rightarrow k \in E$, then $\myfunctionname{ancestor}_T(k,2) = i$.

\item $\myfunctionname{children}_T(i) = \{ k \in V \mid i \rightarrow k \}$,  i.e., the set of children of node $i$,

% For example, given dependencies of the form $h \rightarrow i, \, h \rightarrow g, \, i \rightarrow j, \, i \rightarrow k$, then $\myfunctionname{children}_T(i)$=$\{j,k\}$ and $\myfunctionname{children}_T(h)$=$\{g,i\}$.

\item $\myfunctionname{lm-branch}_T(i,d)= min \{ k \in V \mid i \xrightarrow{d} k \}$, i.e., the set containing the leftmost among the children of $i$ whose dependencies are labeled $d$ (or the empty set if there is no such node).

% For example, given dependencies of the form $i \xrightarrow{d} j, \, i \xrightarrow{d} k$, $\myfunctionname{branch}_T(i,d)$=$\{j\}$.

%\begin{itemize}
%\item $ancestor: V \times \mathbb{Z} \times G \longrightarrow V$, where given a node $i \in V$, it returns a node located at $\delta \in \mathbb{Z}$ levels up in $G$ 
%where there exists a path $[(ancestor_{\delta},d_{i}, ancestor_{\delta-1})$,\break$(ancestor_{\delta-1},d_{i}, ancestor_{\delta-2})$...,$(ancestor_{1}$,\break$d_{i},i)]$
%(e.g. given two dependency triplets: $(i,d,j)$ and $(j,d,k)$, $ancestor(k,2,G)$ returns $i$).

% Let $G=\{(w_{i},d_{(i,j)},w_{j})\}$ be a dependency graph of $S$, where $w_{i}$ and $w_{j}$ are the head
% and the dependent terms respectively, and $d_{(i,j)} \in D $ a dependency type where $D$ is the set of candidate syntactic relationships.
%\item $children: V \times G \longrightarrow V$, where given a node $i \in V$, it returns a set $V_{children} \subseteq V$ where $\forall\ v_{child} \in V_{children}\ \exists\  ancestor(v_{child},1,G)$=$1$ (e.g. given thre triplets $(h,d_{i},i)$, $(h,d_{j},g)$, $(i,d_{j},j)$ and $(i,d_{k},k)$, then $children(i,G)$=$\{j,k\}$ and $children(h,G)$=$\{g,i\}$.
%\item $branch: V \times D \times G \longrightarrow V$,where given a node $i \in V$, an specified branch $d_{branch}$ and $children(i,G)$, it obtains a node $branch \in children(i,G)$
%where the dependency triplet $(i,d_{branch},branch) \in E$ (e.g given two triplets:
%$(i,d1,j)$ and $(i,d2,k)$, $branch(i,d1,G)$ returns $j$).

\end{itemize}

%For practical reasons, we add a fourth element, $\varepsilon_{i}$, to the tuple $(i,d,j) \longrightarrow (\varepsilon,i,d,j) $, to include the semantic orientation at $i$ in an specific step of time.

Our compositional \textsc{sa} system will associate an {\sc so} value $\sigma_{i}$ to each node $i$ in the dependency tree of a sentence, representing the {\sc so} of the subtree rooted at $i$. The system will use a set of compositional operations to propagate changes to the semantic orientations of the nodes in the tree. Once all the relevant operations have been executed, the {\sc so} of the sentence will be stored as $\sigma_0$, i.e., the semantic orientation of the root node.

A compositional operation is triggered when a node in the tree matches a given condition (related to its associated PoS tag, dependency type and/or word form); it is then applied to a scope of one or more nodes calculated from the trigger node by ascending a number of levels in the tree and then applying a scope function. More formally, we define our operations as follows:

\begin{mydef}
Given a dependency tree $T(V,E)$, a \concept{compositional operation} is a tuple $o=(\tau,C,\delta,\pi,S)$ such that:
\begin{itemize}
\itemsep-0.1em
\item $\tau : \mathbb{R} \rightarrow \mathbb{R}$ is a {\bfseries transformation function} to apply on the {\sc so} ($\sigma$) of nodes,
\item $C : V \rightarrow \{true,false\}$ is a predicate that determines whether a node in the tree will \concept{trigger} the operation,
\item \changed{$\delta \in \mathbb{N}$ is a number of levels that we need to ascend in the tree to calculate the scope of $o$, i.e., the nodes of\, $T$ whose SO is affected by the transformation function $\tau$,}
\item $\pi$ is a priority that will be used to break ties when several operations coincide on a given node, and
\item $S$ is a scope calculation function that will be used to determine the nodes affected by the operation.
\end{itemize}
\end{mydef}

In practice, our system defines $C(i)$ by means of sets of words, part-of-speech tags and/or dependency types such that the operation will be triggered if $w_i$, $t_i$ and/or the head dependency of $i$ are in those sets. \changed{Compositional operations where $C(i)$ is defined using only universal tags and dependency types, and which therefore do not depend on any specific words of a given language, can be shared across languages, as showed in \S \ref{section-experiments}.}

% Compositional operations where $C(i)$ is defined using universal tags and dependency types only are universal and can be used across languages. 

% \changed{For the set of of words of $C(i)$, there exists feasible options for m our system can plug any SentiStrength Sendidata \cite{Thelwall:2012}, which provides, among other tables, a list of negating and intensification terms for up to 34 languages.} 
% Additionally, optional language-specific tuning is possible by adding predicates that match specific word forms, tags or language-dependent dependencies.

We propose two options for the transformation function $\tau$:
\begin{itemize}
\itemsep-0.1em
\item $\myfunctionname{shift}_{\alpha}(\sigma) = \left\{
	\begin{array}{ll}
		\sigma - \alpha  & \mbox{if } \sigma > 0 \\
		\sigma + \alpha & \mbox{if } \sigma <0 \\
	\end{array}
\right.$ where $\alpha$ is the shifting factor and $\alpha,\sigma \in \mathbb{R}$.
 \item $\myfunctionname{weighting}_{\beta}(\sigma) = \sigma \times (1+\beta)$ where $\beta$ is the weighting factor and $\beta,\sigma \in \mathbb{R}$.\footnote{\changed{From a theoretical point of view, $\beta$ is not restricted to any value. In a practical implementation, $\beta$ values (which will vary according to the intensifier) should serve to intensify, diminish or even cancel the $\sigma$ of the affected scope in a useful way. In this article, $\beta$'s for intensifiers are directly taken from existing lexical resources and are not tuned in any way, as explained in \S \ref{section-experiments}.}}
\end{itemize}

\begin{figure*}[t]
  \centering
  \includegraphics[width=14.0cm,clip]{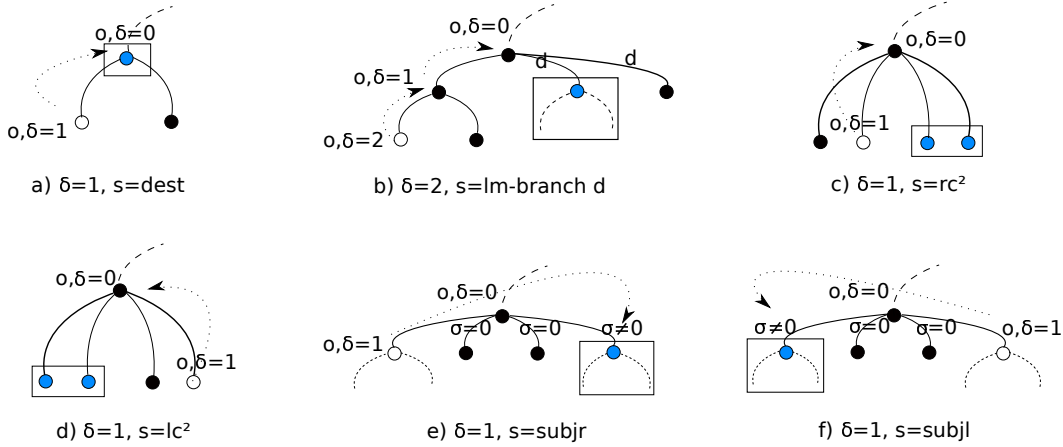}
  \caption{Graphical representation of the proposed set of influence scopes S. $\bigcirc$ indicates the node that triggers an operation $o$, $\square$ the nodes to which it is applied (colored in blue).}
  \label{figure-rules}
\end{figure*}

The scope calculation function, $S$, allows us to calculate the nodes of $T$ whose {\sc so} is affected by the transformation $\tau$. For this purpose, if the operation was triggered by a node $i$, we apply $S$ to $\myfunctionname{ancestor}_T(i,{\delta})$, i.e., the $\delta$th ancestor of $i$ (if it exists), which we call the \concept{destination node} of the operation. The proposed scopes are as follows (see also Figure \ref{figure-rules}):
\begin{itemize}
\itemsep-0.1em
 \item $\myfunctionname{dest}$ (\emph{destination node}): The transformation $\tau$ is applied directly to the {\sc so} of $\myfunctionname{ancestor}_T(i,\delta)$
 (see Figure \ref{figure-rules}.a).
 \item  $\myfunctionname{lm-branch}^d$ (\emph{branch of $d$}): The affected nodes are $\myfunctionname{lm-branch}_T$ $( \myfunctionname{ancestor}_T(i,$ $\delta) ,d)$
 (see Figure \ref{figure-rules}.b).
 \item $\myfunctionname{rc}^n$ (\emph{$n$ right children}): $\tau$ affects the {\sc so} of the $n$ smallest indexes of $\{ j \in \myfunctionname{children}_T( \myfunctionname{ancestor}_T(i,\delta) ) \mid j > i \}$, \changed{i.e., it modifies the global $\sigma$ of the closest (leftmost) $n$ right children of $\myfunctionname{ancestor}_T(i,\delta)$ } (see Figure \ref{figure-rules}.c).
 \item $\myfunctionname{lc}^n$ (\emph{$n$ left children}): The transformation affects the $n$ largest elements of $\{ j \in \myfunctionname{children}_T( \myfunctionname{ancestor}_T(i,\delta) ) \mid j < i \}$, \changed{i.e., it modifies the global $\sigma$ of the closest (rightmost) $n$ left children of $\myfunctionname{ancestor}_T(i,\delta)$ } (see Figure \ref{figure-rules}.d).\footnote{\changed{$\myfunctionname{lc}^n$ and $\myfunctionname{rc}^n$ might be useful in dependency structures where elements such as some coordination forms (e.g. it is \quotedtext{very expensive and bad}) are represented as children of the same node, for example.}}
 \item $\myfunctionname{subjr}$ (\emph{first subjective right branch}): The affected node is $\min \{ j \in \myfunctionname{children}_T$ $( \myfunctionname{ancestor}_T(i,\delta) ) \mid j > i \wedge \sigma_{j} \neq 0 \}$, \changed{i.e., it modifies the $\sigma$ of the closest (leftmost) subjective right child of $\myfunctionname{ancestor}_T(i,\delta)$}  (see Figure \ref{figure-rules}.e).
 \item $\myfunctionname{subjl}$ (\emph{first subjective left branch}): The affected node is $\max \{ j \in \myfunctionname{children}_T$ $( \myfunctionname{ancestor}_T(i,\delta) ) \mid j < i \wedge \sigma_{j} \neq 0 \}$, \changed{i.e., it modifies the $\sigma$ of the closest (rightmost) subjective left child of $\myfunctionname{ancestor}_T(i,\delta)$ }  (see Figure \ref{figure-rules}.f).
\end{itemize}

%It is important to remark that compositional operations are not restricted to any language or dependency annotation criterion. This set of operations can be easily modified to address language-dependent phenomena  or to a treebank following different guidelines by simply defining new operations to add language-specific rules (see \S\ \ref{section-aglorithm-unsupervised}).

Compositional operations can be defined for any language or dependency annotation criterion. 
While it is possible to add rules for language-specific phenomena if needed (see \S\ \ref{section-aglorithm-unsupervised}), in this paper we focus on universal rules to obtain a truly multilingual system. \changed{Apart from universal treebanks and PoS tags, the only extra information used by our rules is a short list of negation words, intensifiers, adversative conjunctions and words introducing conditionals (like the English ``if'' or ``would''). While this information is language-specific, it is standardly included in multilingual sentiment lexica which are available for many languages (\S\ \ref{section-aglorithm-unsupervised}), so it does not prevent our system from working on a wide set of languages without any adaptation, \changedtwo{apart from modifying the subjective lexicon.}}

%This set of operations can be easily modified to address language-dependent phenomena  or to a treebank following different guidelines by simply defining new operations to add language-specific rules (see \S\ \ref{section-aglorithm-unsupervised}).

\subsection{An algorithm for unsupervised SA}\label{section-aglorithm-unsupervised}

\begin{algorithm}[h]
\normalsize{
\caption{Compute SO of a node}
\label{algorithm-compute-node}
\begin{algorithmic}[1]

\small
\Procedure{compute}{$i$, $O$ ,$T$}
\LeftComment{Initialization of queues}
\State $A_{i} \gets []$ 
\State $Q_{i} \gets []$ 

\LeftComment{Enqueue operations triggered by node $i$:}
\For {$o=(\tau,C,\delta,\pi,S)$ in $O$}
\If {$C(i)$}
  \If {$\delta > 0$}
    \State$push((\tau,C,\delta,\pi,S),Q_{i})$
  \Else
    \State $push((\tau,C,\delta,\pi,S),A_{i})$
  \EndIf
\EndIf
\EndFor

\LeftComment{Enqueue operations coming from child nodes:}
%\State $\textit{children} \gets \myfunctionname{children}_T(i)$
\For{$c$ in $\myfunctionname{children}_T(i)$ }
    \For{$o=(\tau,C,\delta,\pi,S)$ in $Q_{c}$}
      %$\delta_{q_{c}} \gets \delta_{q_{c}} -1$ 
      \If {$\delta - 1 = 0$} 
 	  \State $push((\tau,C,\delta - 1,\pi,S),A_{i})$
       \Else
	  \State $push(\tau,C,\delta - 1 ,\pi,S),Q_{i})$
       \EndIf
    \EndFor
\EndFor

\LeftComment{Execute operations that have reached their destination node:}
 \While{$ A_{i}\ is\ not\ empty $}
    \State $o=(\tau,C,\delta,\pi,S) \gets pop(A_{i})$
	\For {$j$ in $S(i)$}
      \State $\sigma_{j} \gets \tau(\sigma_{j})$
	\EndFor
  \EndWhile
\LeftComment{Join the SOs for node \emph{i} and its children:}
\State $\sigma_{i} \gets \sigma_{i} + \sum_{c \in \myfunctionname{children}_T(i)}^{} \sigma_{c}$

\EndProcedure

\end{algorithmic}}
\end{algorithm}

To execute the operations and calculate the {\sc so} of each node in the dependency tree of the sentence, we start by initializing the {\sc so} of each word using a subjective lexicon, in the manner of traditional unsupervised approaches \cite{Turney2002a}. 

Then, we traverse the parse tree in postorder, applying Algorithm \ref{algorithm-compute-node} to update semantic orientations when visiting each node $i$. In this algorithm, $O$ is the set of compositional operations defined in our system, $A_{i}$ is a priority queue of the compositional operations to be applied at node $i$ (because $i$ is their destination node); and $Q_{i}$ is another priority queue of compositional operations to be queued for upper levels at node $i$ (as $i$ is not yet their destination node). $Push$ inserts $o$ in a priority queue and $pop$ pulls the operation with the highest priority (ties are broken by giving preference to the operation that was queued earlier). \changed{When visiting a node, a $push$ into $Q_{i}$ (Algorithm \ref{algorithm-compute-node}, line 7) is executed when the node $i$ triggers an operation $o$ that must be executed at the ancestor of $i$ located $\delta$ levels upward from it. A $push$ into $A_{i}$ (
Algorithm \ref{algorithm-compute-node}, line 9) is executed when the node $i$ triggers an operation that must be executed at that same node $i$ (i.e., $\delta=0$). On the other hand, at node $i$, the algorithm must also decide what to do with the operations coming from $children_T(i)$. Thus, a $push$ into $A_{i}$ (Algorithm \ref{algorithm-compute-node}, line 13) is made when an operation from a child has reached its destination node (i.e., $\delta-1=0$), so that it must be applied at this level. A $push$ into $Q_{i}$ (Algorithm \ref{algorithm-compute-node}, line 15) is made when the operation has still not reached its destination node and must be spread $\delta-1$ more levels up.}

%Let $A_{i}$ be a priority queue of the compositional operations to be applied at node $i$ (because $i$ is their destination node). Let $Q_{i}$ be another priority queue of compositional operations to be queued for upper levels at node $i$ (as $i$ is not yet their destination node).
%Let $Q_{c}$ be the set of queued operations for each $c \in children(i,G)$. Then, Algorithm \ref{algorithm-compute-node} shows the pseudocode to compute $\theta$ at each node, where $\oplus$ defines the operation to merge two priority queues, $push$ inserts $o$ in a priority queue and $pop$ pulls the operation with the highest priority (in case of time, we give preference to the operation that has been queued for longer, i.e., the operation triggered most on the left of $G$).

At a practical level, the set of compositional operations are specified using a simple {\sc xml} file:
\begin{itemize}
\itemsep-0.1em
 \item \texttt{<forms>}: Indicates the tokens to be taken into account for the condition $C$ that triggers the operation. Regular expressions are supported.
 \item \texttt{<dependency>}: Indicates the dependency types taken into account for $C$.
 \item \texttt{<postags>}: Indicates the PoS tags that must match to trigger the rule.
 \item \texttt{<rule>}: Defines the operation to be executed when the rule is triggered.
 \item \texttt{<levelsup>}: Defines the number of levels from $i$ to spread before applying $o$.
 \item \texttt{<priority>}: Defines the priority of $o$ when more than one operation needs to be applied over $i$ (a larger number implies a bigger priority).
\end{itemize}

\subsection{NLP tools for universal unsupervised SA}\label{section-nlp-tools}

The following resources serve us as the starting point to carry out state-of-the-art universal, unsupervised and syntactic sentiment analysis.

The system developed by \newcite{gimpel2011part} is used for tokenizing. Although initially intended for English tweets, we have observed that it also performs robustly for many other language families (Romance, Slavic, etc.). 
For part-of-speech tagging we rely on the free distribution of the \newcite{TouMan2000a} tagger. Dependency parsers are built using MaltParser \cite{MaltParser} and MaltOptimizer \cite{BalNiv2012}. We trained a set of taggers and parsers for different languages using the universal tag and dependency sets \cite{PetDasMcD2011a,McdNivQuirGolDasGanHalPetZhaTacBedCasLee2013}. \changed{In particular, we are relying on the monolingual models using universal part-of-speech tags presented by \newcite{VilGomAloACL2016}.}

\changed{With respect to multilingual lexical resources, there are a number of alternatives: SentiStrength (subjective data for up to 34 languages); the \newcite{CheSki2014a} approach, which introduced a method for building sentiment lexicons for 136 languages; or SentiWordNet \cite{esuli2006sentiwordnet}, where each synset from WordNet is assigned a objective, positive and negative score.}
Our implementation supports the lexicon format of SentiStrength, which can be plugged directly into the system. Additionally, we provide the option to create different dictionary entries depending on PoS tags to avoid conflicts between homonymous words (e.g. \emph{\textquoteleft I'm fine\textquoteright} versus \emph{\textquoteleft They gave me a fine\textquoteright}).

\section{Defining compositional operations}\label{section-real-compositional-operations}

%We showed above how it is possible to mathematically define arbitrarily complex compositional operations for unsupervised {\sc sa} over a dependency tree.
We presented above a formalism to define arbitrarily complex compositional operations for unsupervised {\sc sa} over a dependency tree.
In this section, we show the definition of the most important rules that we used to evaluate our system.
In practical terms, this implies studying how syntactic constructions that modify the sentiment of an expression are represented in the annotation formalism used for the training of the dependency parser, in this case, Universal Treebanks.
We are using examples following those universal guidelines, since they are available for more than 40 languages and, as shown in \S\ \ref{section-experiments}, the same rules can be competitive across different languages.

\subsection{Intensification}

Intensification amplifies or diminishes the sentiment of a word or phrase. Simple cases of this phenomenon can be \emph{\textquoteleft I have {\bfseries huge} problems\textquoteright} or \emph{\textquoteleft This is {\bfseries a bit} dissapointing\textquoteright}. Traditional lexicon-based methods handle most of these cases with simple heuristics (e.g. amplifying or diminishing the sentiment of the word following an intensifier). 
However, ambiguous cases might appear where such lexical heuristics are not sufficient. For example, \emph{\textquoteleft huge\textquoteright} can be a subjective adjective introducing its own {\sc so} (e.g. \emph{\textquoteleft The house is huge\textquoteright}), but also an amplifier when it modifies a subjective noun or adjective (e.g. \emph{\textquoteleft I have huge problems\textquoteright}, where it makes \emph{\textquoteleft problems\textquoteright} more negative). 

% About this issue, for example, \newcite{Lexicon-BasedMethods} say: \emph{\textquoteleft When an intensifying adjective appears next to an SO-valued noun, it is treated as an intensifier rather than as a separate SO-bearing
% unit.\textquoteright}. The concept of \textquotedblleft next to\textquotedblright\ is limited from a lexical point of view: the level of proximity can be dependent of the language or the \textcolor{red}{intensifier might even appear before the intensified term}.\changed{ For example, \emph{\textquoteleft This is damn good\textquoteright} could be computed as -4 + 3 (damn as a negative adjective and good as an adjective) or 1.35*3 (damn as an intensifier and good as an adjective).\footnote{Examples computed following \newcite{Lexicon-BasedMethods} lexica.}}

 \begin{figure}[thbp]
  \centering
  \includegraphics[width=7.5cm,clip]{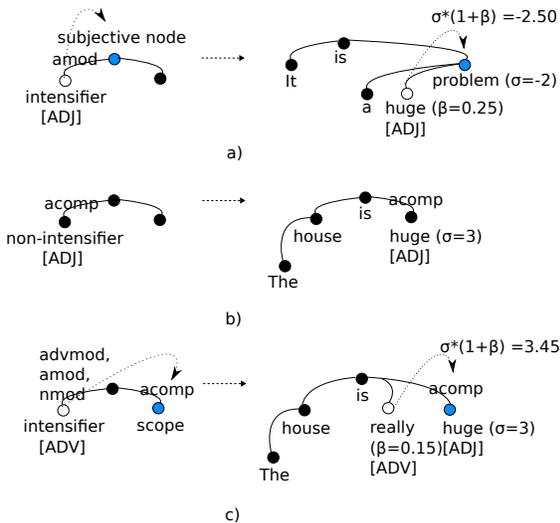}
  \caption{Skeleton for intensification compositional operations (2.a, 2.c) and one case without intensification (2.b), together with examples annotated with universal dependencies. \changed{Semantic orientation values are for instructional purposes only. In 2.a, \quotedtext{huge} is a term considered in a list of intensifiers, labeled as an ADJ, whose dependency type is \emph{amod}, matching the definition of the intensification compositional operation. As a result, the $o$ for intensification is triggered, spreading $\delta=1$ levels up (i.e., up to \quotedtext{problem}) and amplifying the $\sigma$ of \emph{dest} (the first scope of the operation that matches, i.e., \quotedtext{problem}) by (1+$\beta$). In 2.b,  \quotedtext{huge} is again a word occurring in the intensifier list and tagged as an ADJ, but its dependency type is \emph{acomp}, which is not considered among the intensification dependency types. As a result, no operation is triggered and the word is treated as a regular word (introducing its 
own {\sc so} rather than modifying others). In 2.c, \quotedtext{really} is the term 
acting as intensifier, triggering again an intensification operation on the node $\delta=1$ levels up from it (\quotedtext{is} node). Differently from 2.a, in this case the scope $dest$ is not applicable since the word \quotedtext{is} is not subjective, but there is a matching for the second candidate scope, the branch labeled as $acomp$ (the branch rooted at \quotedtext{huge}), so the $\sigma$ associated with that node of the tree is amplified.}}
  \label{figure-intensifier-rules}
\end{figure}

Universal compositional operations overcome this problem without the need of any heuristic. A dependency tree already shows the behavior of a word within a sentence thanks to its dependency type, and it shows the role of a word independently of the language. Figure \ref{figure-intensifier-rules} shows graphically how universal dependencies represent the cases discussed above these lines. Formally, the operation for these forms of intensification is:
$(\myfunctionname{weighting}_{\beta}, {w \in {\text{\small{intensifiers}}}} \wedge {t \in \{\text{\small{ADV,ADJ}}\}} \wedge {d \in \{\text{\small{advmod,amod,nmod}}\}} , 1, 3, $ ${\myfunctionname{dest} \cup \myfunctionname{lm-branch}^{\text{acomp}}})$, with the value of $\beta$ depending on the strength of the intensifier as given by the sentiment lexicon.

\subsubsection{\emph{\textquoteleft But\textquoteright} clauses}

%  \begin{figure}[hbtp]
%   \centering
%   \includegraphics[width=4.0cm,clip]{intensifier-non-intensifier}
%   \caption{\textcolor{red}{Dependency strucuture under \newcite{McdNivQuirGolDasGanHalPetZhaTacBedCasLee2013} universal guidelines where intensfier adjectives are labeled differently from non-intensifer adjectives (dependency type is acomp).}}
%   \label{figure-intensifier-acting-as-non-intensfier}
% \end{figure}

\begin{figure}[hbtp]
  \centering
  \includegraphics[width=7.5cm,clip]{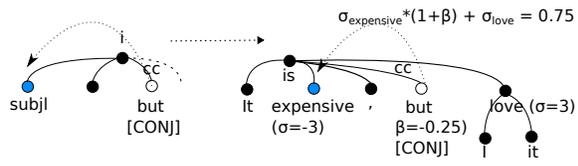}
  \caption{Skeleton for \emph{\textquoteleft but\textquoteright} compositional operation illustrated with one example according to universal dependencies. \changed{The term \quotedtext{but} matches the word form, tag and dependency types required to act as a sentence intensifier, so the compositional operation is queued to be applied $\delta=1$ levels upward (i.e., at the \quotedtext{is} node). The scope of the operation is the first subjective branch that is a left child of said \quotedtext{is} node (i.e., the branch rooted at \quotedtext{expensive}). As a result, the $\sigma$ rooted at this branch is diminished by multiplying it by (1+$\beta$) (note that $\beta$ is negative in this case) and the resulting value is added to the $\sigma$ computed at \quotedtext{is} for the rest of the subjective children.}}
  \label{figure-but-rules}
\end{figure}

Compositional operations can also be defined to manage more challenging cases, such as clauses introduced by \emph{\textquoteleft but\textquoteright}, considered as a special case of intensification by authors such as \newcite{BroTofTab2009a} or \newcite{VilAloGom2015a}. It is assumed that the main clause connected by \emph{\textquoteleft but\textquoteright} becomes less relevant for the reader (e.g. \emph{\textquoteleft It is expensive, {\bfseries but} I love it\textquoteright}). Figure \ref{figure-but-rules} shows our proposed composition operation for this clause, formally: $( \myfunctionname{weighting}_{\beta}, w \in \{\text{\small{but}}\}\ \wedge t \in \{\text{\small{CONJ}}\} \wedge d \in \{\text{\small{cc}}\} , 1, 1, \myfunctionname{subjl})$ with $\beta = -0.25$. Note that the priority of this operation ($\pi=1$) is lower than that of intensification ($\pi=3$), since we first need to process intensifiers, which are local phenomena, before resolving adversatives, which have a larger scope.

% \textcolor{red}{Note that differently from existing language-dependent syntactic approaches \cite{VilAloGom2015a}, compositional operations easily allow to implement
% different ways of computing clauses like this.} 

\subsection{Negation}

\begin{figure}[hbtp]
  \centering
  \includegraphics[width=7.5cm,clip]{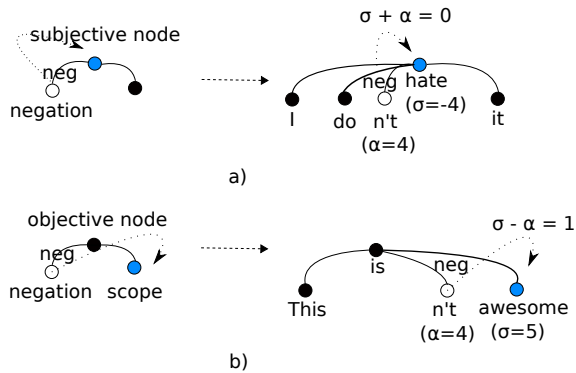}
  \caption{Skeleton for negation compositional operations illustrated together with one example. \changed{In 5.a, the term \quotedtext{n't} matches the form word of a negator and its dependency type is \emph{neg}, queuing a negation compositional operation to be applied $\delta=1$ levels upward (i.e., at the \quotedtext{hate} node). The first candidate scope for that operation matches, because $dest$ is a subjective word (\quotedtext{hate}), shifting the $\sigma$ of such word according to the definition of our $shift_\alpha(\sigma)$ transformation function. In a similar way, in 5.b, \quotedtext{n't} also acts a negator term, but in this case the candidate scope that matches is the second one (i.e.,   $\myfunctionname{lm-branch}^{\text{attr}}$ ).}}
  \label{figure-negation-rules}
\end{figure}

Negation is one of the most challenging phenomena to handle in {\sc sa}, since its semantic scope can be non-local
(e.g. \emph{\textquoteleft I do not plan to make you suffer\textquoteright}).
Existing unsupervised lexical approaches are limited to considering a snippet to guess the scope of negation. Thus, it is likely that they consider as a part of the scope terms that should not be negated from a semantic point of view.
Dependency types
help us to determine which nodes should act as negation and which should be its scope of influence. For brevity, we only illustrate some relevant negation cases and instructional examples in Figure \ref{figure-negation-rules}.
Formally, the proposed compositional operation to tackle most forms of negation under universal guidelines is: $( \myfunctionname{shift}_{\alpha}, w \in \text{negations} \wedge t \in U \wedge d \in \{\text{neg}\} , 1, 2, \myfunctionname{dest}\ \cup\ \myfunctionname{lm-branch}^{\text{attr}}\ \cup\ \myfunctionname{lm-branch}^{\text{acomp}}\ \cup\ \myfunctionname{subjr})$, where $U$ represents the universal tag set. The priority of negation ($\pi=2$) is between those of intensification and \emph{\textquoteleft but\textquoteright} clauses because its scope can be non-local, but it does not go beyond an adversative conjuction.

\subsection{Irrealis}

\emph{Irrealis} denotes linguistic phenomena used to refer to non-factual actions, such as conditional, subjunctive or desiderative sentences (e.g. \emph{\textquoteleft He {\bfseries would} have died {\bfseries if} he hadn't gone to the doctor\textquoteright}).
It is a very complex phenomenon to deal with, and systems are either usually unable to tackle this issue or simply define rules to ignore
sentences containing a list of irrealis stop-words \cite{Lexicon-BasedMethods}.
We do not address this phenomenon in detail in this study, but only propose a rule to deal with \emph{\textquoteleft if\textquoteright} constructions
(e.g. \emph{\textquoteleft if I die [...]\textquoteright} or \emph{\textquoteleft if you are happy [...]\textquoteright}, considering that the phrase that contains it should be ignored from the final
computation. Formally: 
$( \myfunctionname{weighting}_{\beta}, w \in \{\text{\small{if}}\} \wedge t \in U \wedge d \in \{\text{\small{mark}}\} , 2, 3, \myfunctionname{dest} \cup \myfunctionname{subjr})$. Its graphical representation would be very similar to intensification (see Figures \ref{figure-rules} a) and e)).

\subsection{Discussion}

\begin{figure*}[hbpt]
   \centering
  \includegraphics[width=13.5cm,clip]{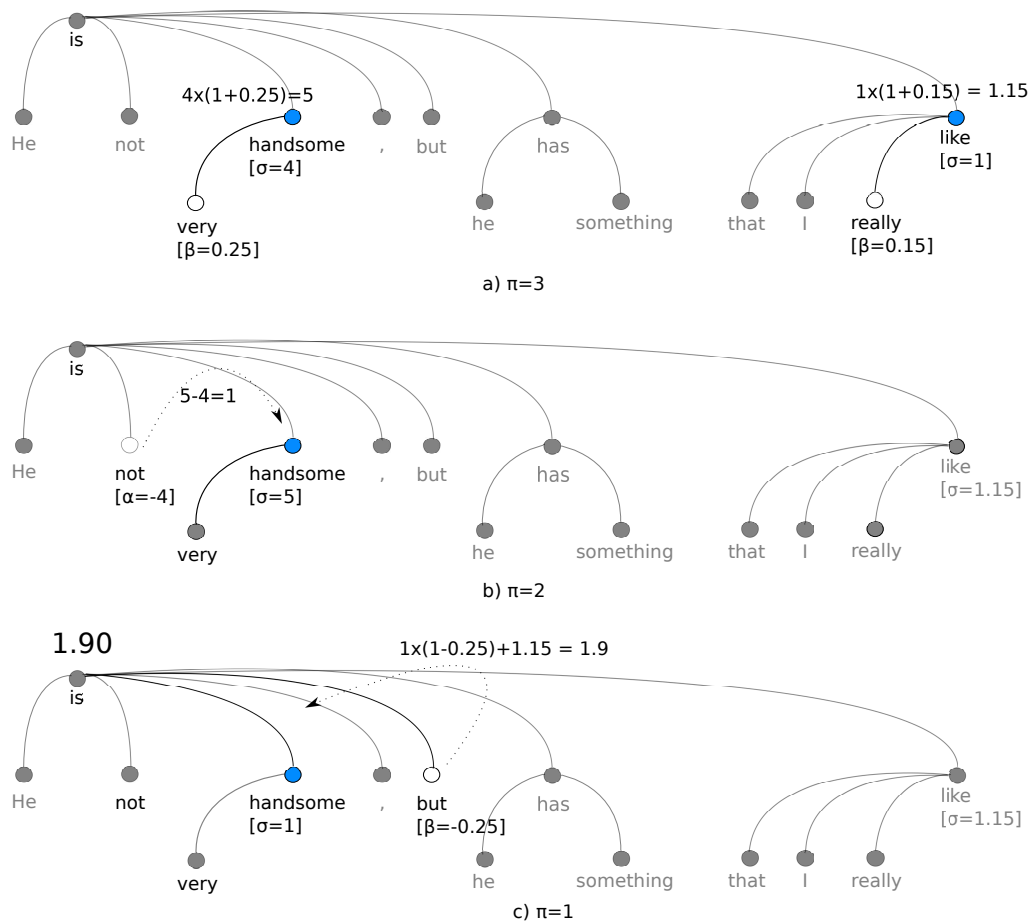}
  \caption{Analysis of a sentence applying universal unsupervised prediction. For the sake of clarity, the real post-order traversal is not illustrated. Instead we show an (in this case) equivalent computation by applying all operations with a given priority, $\pi$, at the same time, irrespective of the node. Semantic orientation, intensification and negation values are extracted from the dictionaries of \newcite{Lexicon-BasedMethods}. Phase a) shows how the intensification is computed on the branches rooted at \emph{\textquoteleft handsome\textquoteright} and \emph{\textquoteleft like\textquoteright}. Phase b) shows how the negation shifts the semantic orientation of the attribute (again, the branch rooted at \emph{\textquoteleft handsome\textquoteright}). Phase c) illustrates how the clause \emph{\textquoteleft but\textquoteright} diminishes the semantic orientation of the main sentence, in particular the semantic orientation of the attribute, the first left subjective branch of its head. Elements 
that are not playing a role in a specific phase appear dimmed. One of the interesting points in this example comes from illustrating how three different phenomena involving the same branch (the attribute \emph{\textquoteleft handsome\textquoteright}) are addressed properly thanks to the assigned $\pi$.}
  \label{figure-running-example}
\end{figure*}

\changed{Figure \ref{figure-running-example} represents an analysis of our introductory sentence \emph{\textquoteleft He is not very handsome, but he has something that I really like\textquoteright}, showing how compositional operations accurately capture semantic composition.\footnote{\changed{The system released together with this paper shows an equivalent ASCII text representation that can be obtained on the command line. It is also possible to check how the system works at https://miopia.grupolys.org/demo/}}} \changed{Additionally, Table \ref{table-algorithm-state-by-step} illustrates the internal state of the algorithm and the {\sc so} updates made at each step for our running example.}

\begin{table}[hbtp]
\begin{center}
\tiny{
\tabcolsep=0.020cm
\begin{tabular}{|cccccc|}
\hline
\bf Step&\bf Word$_{index}$&\bf A$_{word(\delta,\pi)}$&\bf Q$_{word(\delta,\pi)}$&\bf $\sigma_{word}$&\bf $\sigma_{word} \leftarrow A$\\
\hline
\hline
1&He$_1$&[\,]&[\,]&0&0\\
2&not$_3$&[\,]&[N$_{not(1,2)}$]&0&0\\
3&very$_4$&[\,]&[I$_{very(1,3)}$]&0&0\\
4&handsome$_5$&[I$_{very(0,3)}$]&[\,]&4&5\\
5&,$_6$&[\,]&[\,]&0&0\\
5&but$_7$&[\,]&[I$_{but(1,1)}$]&0&0\\
6&he$_8$&[\,]&[\,]&0&0\\
7&something$_{10}$&[\,]&[\,]&0&0\\
8&has$_9$&[\,]&[\,]&0&0\\
9&I$_{12}$&[\,]&[\,]&0&0\\
10&that$_{11}$&[\,]&[\,]&0&0\\
11&really$_{12}$&[\,]&[I$_{really(1,3)}$]&0&0\\
12&like$_{13}$&[I$_{really(1,3)}$]&[\,]&1&1.15\\
13&is$_2$&[N$_{not(0,2)}$, I$_{but(0,1)}$]&[\,]&0&1.90\\
\hline
\end{tabular}}
\end{center}
\caption{\changed{Internal state and {\sc so} updates made by the proposed algorithm for the running example. Each row corresponds to a step in which a node (Word$_{index}$) is visited in the postorder traversal. Columns A$_{word(\delta,\pi)}$ and Q$_{word(\delta,\pi)}$ show the state of the queues after the enqueuing operations, but before $A$ is emptied (i.e., immediately before line 16 of Algorithm \ref{algorithm-compute-node}). The $\sigma_{word}$ column shows the {\sc so} of the visited node at that same point in time, and $\sigma_{word} \leftarrow A$ is the new {\sc so} that is assigned by applying compositional operations and joining the {\sc so}s of children (lines 16-20 of Algorithm \ref{algorithm-compute-node}). \emph{N} and \emph{I} refer to negation and intensification operations.}}
\label{table-algorithm-state-by-step}
\end{table}

It is hard to measure the coverage of our rules and the potential of these universal compositional operations, since it is possible to define arbitrarily complex operations for as many relevant linguistic phenomena as wished. In this line, \newcite{PorCamWinHua2014a} define a set of English sentic patterns to determine how sentiment flows from concept to concept in a variety of situations (e.g. relations of complementation, direct nominal objects, relative clauses, \dots) over a dependency tree following the \newcite{de2008stanford} guidelines. \changed{The main difference of our work with respect to \newcite{PorCamWinHua2014a} or \newcite{VilAloGom2015a} is that they present predefined sets of linguistic patterns for language-specific SA, whereas our approach is a theoretical formalism to define arbitrarily complex patterns given tagging and parsing guidelines, which has been implemented and tested on a universal set of syntactic annotation guidelines that work across different languages (see \S \ref{section-experiments}).}

\changed{
Under this approach, switching the system from one language to another only requires having a tagger and a parser following the Universal Treebanks (v2.0) guidelines and a subjectivity lexicon, but compositional operations remain unchanged (as shown in \S 5).\footnote{\changed{There is a difference between the number of compositional operations that are defined in the system (one for each phenomenon considered: intensification, \quotedtext{but} clauses, negation and irrealis), and the number of compositional operation instances created at runtime given such definitions. The latter depends on the words, tags and dependency types that match each operation's predicate $C$. While the matching tags and dependency types are fixed and common to all languages, the number of words that can match $C$ depends on the lexicon, so the number of operation instances varies across languages depending on the use of SO-CAL and SentiStrength as lexical resources (e.g. English is the language that generates more instances, with 
\changed{1411 compositional operations}, due to having the largest intensifier lexicon among the languages and resources considered)}.}}

\changed{The performance of the algorithm might vary according to the quality of the resources on which it relies. Mistakes committed by the tagger 
and the parser might have some influence on the approach. However, preliminary experiments on English texts show that having a parser with a LAS\footnote{\changed{Labeled Attachment Score (LAS): The percentage of dependencies where both the head and the dependency type have been assigned correctly. The English model used has a LAS of 89.36\%.}} over 75\% is enough to properly exploit compositional operations. With respect to the lexicalized parsing (and tagging) models, usually a different model is needed per language, even when using universal guidelines. In this respect, recent studies \cite{VilGomAloACL2016,ammar2016one,guo2016exploiting} have showed how it is possible to train a single model on universal treebanks to parse different languages with state-of-the-art results. This makes it possible to universalize one of the most relevant previous steps of our approach. The same steps can be taken to train multilingual tagging models \cite{VilGomAloACL2016}.}

\changed{Adapting or creating new compositional operations for other tagging and parsing guidelines different from Universal Treebanks only requires: (1) becoming familiar with the new tag and dependency sets to determine which tags and dependency types should be included in each $C$, and (2) manually inspecting sentences parsed with the target guidelines to detect if they give a different structural representation of relevant phenomena. In this case, a new set of $S$, $\pi$ or $\delta$ values may be needed, so that we can correctly traverse the tree and determine scopes on such dependency structure. At the moment, new practical operations need to be added manually, by defining them in the {\sc xml} file.}

% We plan to incorporate these patterns as compositional operations as a short-term aim, so they can be handled universally.

\section{Experimental results}\label{section-experiments}

% \changed{We compare our algorithm with respect to both unsupervised and supervised approaches on three languages (English, Spanish and German).}
We compare our algorithm with respect to existing approaches on three languages: English, Spanish and German.
% The choice of languages is due to the availability of annotated sentiment corpora for evaluation and of other unsupervised {\sc sa} systems for comparison.
The availability of corpora and other unsupervised {\sc sa} systems for English and Spanish enables us to perform a richer comparison than in the case of German, where we only have an \emph{ad-hoc} corpus. 
% In the case of German, we use an \emph{ad-hoc} corpus  

We compare our algorithm with respect to two of the most popular and widely used unsupervised systems: (1)  {\sc so-cal} \cite{Lexicon-BasedMethods}, a language-dependent system available for English and Spanish guided by lexical rules at the morphological level, and (2) SentiStrength, a multilingual system that does not apply any PoS tagging or parsing step in order to be able to do multilingual analysis, relying instead on a set of subjectivity lexica, snippet-based rules and treatment of non-grammatical phenomena (e.g. character replication). 
Additionally, for the Spanish evaluation, we also took into account the system developed by \newcite{VilAloGom2015a}, an unsupervised syntax-based approach available for Spanish but, in contrast to ours, heavily language-dependent.

For comparison against state-of-the-art supervised approaches, we consider the deep recursive neural network presented by \newcite{SocPerWuChuManNgPot2013a}, trained on a movie sentiment treebank (English). To the best of our knowledge, there are no semantic compositional supervised methods for Spanish and German.

Accuracy is used as the evaluation metric for two reasons: (1) it is adequate for measuring the performance of classifiers when the chosen corpora are balanced and (2) the selected systems for comparison also report their results using this metric.

\subsection{Resources}\label{section-resources}
We selected the following standard English corpora for evaluation:
\begin{itemize}
\itemsep-0.1em
 \item \newcite{taboada2004analyzing} corpus: A general-domain collection of 400 long reviews (200 positive, 200 negative) about hotels, movies, computers or music among other topics, extracted from epinions.com. 
 \item Pang and Lee 2004 corpus \cite{PangBoandLee2004}: A corpus of 2\,000 long movie reviews (1\,000 positive, 1\,000 negative). 
 \item Pang and Lee 2005 corpus \cite{PanLee2005}: A corpus of short movie reviews (sentences). In particular, we used the test split used by \newcite{SocPerWuChuManNgPot2013a}, removing the neutral ones, as they did, for the binary classification task (total: 1\,821 subjective sentences).
\end{itemize}

To show the universal capabilities of our system we include an evaluation for Spanish using the corpus presented by \newcite{BroTofTab2009a} (200 positive and 200 negative long reviews from ciao.es). For German, we rely on a dataset of 2\,000 reviews (1\,000 positive and 1\,000 negative reviews) extracted from Amazon.

\changed{As subjectivity lexica, we use the same dictionaries used by {\sc so-cal} for both English (2\,252 adjectives, 1\,142 nouns, 903 verbs, 745 adverbs and 177 intensifiers) and Spanish (2\,049 adjectives, 1\,333 nouns, 739 verbs, 594 adverbs and 165 intensifiers). For German, we use
the German SentiStrength dictionaries \cite{momtazi2012fine} instead (2\,677 stems and 39 intensifiers), as \newcite{BroTofTab2009a} dictionaries are not available for languages other than Spanish or English. \changed{These are freely available resources that avoid the need to collect subjective words, intensifiers or negators. We just take those resources and directly plug them into our system. The weights were not tuned or changed in any way.\footnote{\changed{To test the soundness of our theoretical formalism and the practical viability and competitiveness of its implementation, it does not matter what resource is chosen. We could have selected other available lexical resources such as SentiWordNet. The motivation for choosing SentiStrength (and SO-CAL) dictionaries is purely evaluative. We have compared our model with respect to other three state-of-the-art and widely used SA systems that use said resources. Our aim is not to evaluate our algorithm over a variety of different lexical resources, but to 
check if our universal system and compositional operations can compete with 
existing 
unsupervised systems under the same conditions (namely, using the same dictionaries and analogous sets of rules).}}} The list of emoticons from Sentistrength is also used as a lexical resource. If a term does not appear in these dictionaries, it will not have 
any impact on the computation of the {\sc so}.\footnote{\changed{Out-of-vocabulary words are not given a special treatment at the moment.}} The content of these dictionaries and their parameters are not modified or tuned.}

\subsection{Comparison to unsupervised approaches}\label{section-experiments-unsupervised-approaches}

Table \ref{table-unsupervised-english} compares the performance of our model with respect to SentiStrength\footnote{We used the default configuration, which already applies many optimizations. We set the length of the snippet between a negator and its scope to 3, based on empirical evaluation, and applied the configuration to compute sentiment on long reviews.}  and {\sc so-cal} on the \newcite{taboada2004analyzing} corpus. 
% Despite of using the same dictionaries, our baseline is a bit behind from \newcite{Lexicon-BasedMethods}, since the authors applied some lemmatization stripping strategy, that was not considered by us, since those are dependent of the language.\footnote{Our system provides the option to plug a lemmas lookup table, but it was not used for the evaluation in English in order to get a fair comparison.} 
% \textcolor{red}{It is worth noting that while {\sc so-cal} and SentiStrength take into account non-grammatical phenomena such as capitalization or character repetition, we do not address them, as it is out of the scope of this study.
% %In spite of this, our algorithm beats both rule-based systems under all configurations, except for the baseline. With respect to {\sc so-cal}, results show that: (1) we handle better negation and intensification (overall improvement of 4.50 percent points) and \textcolor{red}{(2) that we achieve an small improvement when including our treatment irrealis, but there is space for improvement if we compare it with respect to the {\sc so-cal}}. With respect SentiStrength, our system seems to perform better on long reviews.
% In spite of this, our algorithm outperforms both rule-based systems under all configurations, except for the baseline (where no compositional rules are applied).}
With respect to {\sc so-cal}, results show that our handling of negation and intensification provides better results (outperforming {\sc so-cal} by 3.25 percentage points overall). With respect to SentiStrength, our system achieves better performance on long reviews.

Table \ref{table-unsupervised-english-movies} compares these three unsupervised systems on the Pang and Lee 2004 corpus \cite{PangBoandLee2004}, showing the robustness of our approach across different domains. Our system again performs better than {\sc so-cal} for negation and intensification (although it does not behave as well when dealing with irrealis, probably due to the need for more complex compositional operations to handle this phenomenon), and also better than SentiStrength on long movie reviews.

\begin{table}[hbtp]
\begin{center}
\small{
\tabcolsep=0.090cm
\begin{tabular}{|lccc|}
\hline
Rules &\bf SentiStrength &\bf SO-CAL &\bf Our system \\
\hline
\hline
Baseline         &N/A     &\bf 65.50 & 65.00\\
+negation        &N/A     & 67.75 &\bf 71.75\\
+intensification &66.00 & 69.25& \bf 74.25\\
+irrealis        &N/A     & 71.00& \bf 73.75 \\
\hline
\end{tabular}}
\end{center}
\caption{Accuracy (\%) on the \newcite{taboada2004analyzing} corpus. We only provide one row for SentiStrength since we are using
the standard configuration for English (which already includes negation and intensification functionalities).}
\label{table-unsupervised-english}
\end{table}

\begin{table}[hbtp]
\begin{center}
\small{
\tabcolsep=0.090cm
\begin{tabular}{|lccc|}
\hline
Rules &\bf SentiStrength &\bf SO-CAL & \bf Our system \\
\hline
\hline
Baseline         & N/A &\bf 68.05& 67.77\\
+negation        & N/A & 70.10& \bf 71.85\\
+intensification & 56.90 & 73.47 &\bf 74.00\\
+irrealis        & N/A &\bf 74.95& 74.10\\
\hline
\end{tabular}}
\end{center}
\caption{Accuracy (\%) on Pang and Lee 2004 test set \cite{PangBoandLee2004}.}
\label{table-unsupervised-english-movies}
\end{table}

Table \ref{table-unsupervised-spanish} compares the performance of our universal approach on a different language (Spanish) with respect to: Spanish SentiStrength \cite{VilTheAlo2015a}, the Spanish {\sc so-cal} \cite{BroTofTab2009a} and a syntactic language-dependent system inspired on the latter \cite{VilAloGom2015a}. We used exactly the same set of compositional operations as used for English (only changing the list of word forms for negation, intensification and \emph{\textquoteleft  but\textquoteright} clauses, as explained in \S \ref{section-operations-compositional-sa}). Our universal system again outperforms SentiStrength and {\sc so-cal} in its Spanish version.
The system also obtains results very similar to the ones reported by \newcite{VilAloGom2015a}, even though their system is language-dependent and the set of rules is fixed and written specifically for Spanish.

%2004 74,1 + irrealis
%SFU 73.75 +irrealis

%SFUes 75.75 + irrealis, 74.25 +int, 71+neg

\begin{table}[t]
\begin{center}
\tiny{
\tabcolsep=0.090cm
\begin{tabular}{|lcccc|}
\hline
Rules & \bf SentiStrength &\bf SO-CAL & \bf Our system &\bf \newcite{VilAloGom2015a} \\
\hline
\hline
Baseline& N/A & N/A &\bf 63.00&61.80\\
+negation& N/A & N/A &\bf 71.00& N/A\\
+intensification& 73.00 & N/A &74.25&\bf 75.75\\
+irrealis& N/A & 74.50&\bf 75.75& N/A\\
\hline
\end{tabular}}
\end{center}
\caption{Accuracy (\%) on the Spanish \newcite{BroTofTab2009a} test set.}
\label{table-unsupervised-spanish}
\end{table}

% \changed{Table \ref{table-unsupervised-portuguese} compares our model with respect to SentiStrength the performance ({\sc so-cal} is not availability for Portuguese).}
% 
% 
% \begin{table}[t]
% \begin{center}
% \small{
% \tabcolsep=0.050cm
% \begin{tabular}{|lcc|}
% \hline
% Rules &\bf SentiStrength & \bf Our system \\
% \hline
% \hline
% Baseline & N/A &N/A\\
% +negation & N/A &N/A\\
% +intensification& N/A &N/A\\
% +irrealis & N/A&N/A\\
% \hline
% \end{tabular}}
% \end{center}
% \caption{Accuracy (\%) on the Buscape corpus extracted from \newcite{avancco2016normalizaccao}.}
% \label{table-unsupervised-portuguese}
% \end{table}

In order to check the validity of our approach for languages other than  English and Spanish, we have considered the case of German. It is worth noting that the authors of this article have no notions of German at all. In spite of this, we have been able to create a state-of-the-art unsupervised {\sc sa} system by integrating an existing sentiment lexicon into the framework that we propose in this article.

We use the German SentiStrength system \cite{momtazi2012fine} for comparison. The use of the German SentiStrength dictionary, as mentioned in Section \ref{section-resources}, allows us to show how our system is robust when using different lexica. Experimental results show an accuracy of 72.75\% on the Amazon review dataset when all rules are included, while SentiStrength reports 69.95\%. Again, adding first negation (72.05\%) and then intensification (72.85\%) as compositional operations produced relevant improvements over our baseline (69.85\%). The results are comparable to those obtained for other languages, using a dataset of comparable size, reinforcing the robustness of our approach across different domains, languages, and base dictionaries.

\subsection{Comparison to supervised approaches}

Supervised systems are usually unbeatable on the test portion of the corpus with which they have been trained. However, in real applications, a sufficiently large training corpus matching the target texts in terms of genre, style, length, etc. is often not available; and the performance of supervised systems has proven controversial on domain transfer applications \cite{Aue2005}. 

Table \ref{table-supervised-models} compares our universal unsupervised system to \newcite{SocPerWuChuManNgPot2013a} on a number of corpora: (1) the collection used in the evaluation of the Socher et al. system \cite{PanLee2005}, (2) a corpus of the same domain, i.e., movies \cite{PangBoandLee2004}, and (3) the \newcite{taboada2004analyzing} collection.
Socher et al.'s system provides sentence-level polarity classification with five possible outputs: \emph{very positive}, \emph{positive}, \emph{neutral},
\emph{negative}, \emph{very negative}. Since the \newcite{PangBoandLee2004} 
and \newcite{taboada2004analyzing} corpora are collections of long reviews, we needed to collect the global
sentiment of the text. \changed{For the document-level corpora, we count the number of outputs of each class\footnote{\changed{When trying to analyze the document-level corpora with Socher et al.'s system, we had \emph{out-of-memory problems} on a 64-bit Ubuntu server with 128GB of RAM memory, so we decided to choose a counting approach instead over the sentences of such corpora.}} (\emph{very positive} and \emph{very negative} count double, \emph{positive} and \emph{negative} count one and \emph{neutral} counts zero)}. We take the majority class, and in the case of a tie, it is classified as negative.\footnote{These criteria were selected empirically. Assigning the positive class in the case of a tie was also tested, as well as not doubling the \emph{very positive} and \emph{very negative} output, but these settings produced similar or worse results with the \cite{SocPerWuChuManNgPot2013a} system.}

The experimental results show that our approach obtains better results on corpora (2) and (3). It is worth mentioning that our unsupervised compositional approach outperformed the supervised model not only on an out-of-domain corpus, but also on another dataset of the same domain (movies) as the one where the neural network was trained and evaluated. This reinforces the usefulness of an unsupervised approach for applications that need to analyze a number of texts coming from different domains, styles or dates, but there is a lack of labeled data to train supervised classifiers for all of them. As expected, \newcite{SocPerWuChuManNgPot2013a} is unbeatable for an unsupervised approach on the test set of the corpus where it was trained. However, our unsupervised algorithm also performs very robustly on this dataset.

\begin{table}[t]
\begin{center}
\tiny{
\begin{tabular}{|lcc|}
\hline
\bf Corpora &\bf \newcite{SocPerWuChuManNgPot2013a} & \bf Our system \\
\hline
\multicolumn{3}{l}{\emph {Origin corpus of \newcite{SocPerWuChuManNgPot2013a} model}} \\
\hline
Pang and Lee 2005 \cite{PanLee2005} &\bf85.40&75.07\\
\hline
\multicolumn{3}{l}{\emph{Other corpora}} \\
\hline
\newcite{taboada2004analyzing} & 62.00&\bf73.75 \\
Pang and Lee 2004 \cite{PangBoandLee2004} & 63.80&\bf74.10\\
\hline
\end{tabular}}
\end{center}
\caption{Accuracy (\%) on different corpora for \newcite{SocPerWuChuManNgPot2013a} and our system.
On the Pang and Lee 2005 \cite{PanLee2005} collection, our detailed results taking into account different compositional operations were:
73.75 (baseline), 74.13 (+negation), 74.68 (+intensification) and 75.07 (+irrealis)}
\label{table-supervised-models}
\end{table}

%2005 75.01 + irrealis, 74.68 +int 74.13 + neg, 73.75 baseline

% \changed{For Spanish and Portuguese we compare our approach with respect to the supervised models used for comparison in \cite{BroTofTab2009a} and \cite{avancco2016normalizaccao}, respectively. For Spanish, our}
% 
% 
% \begin{table}[t]
% \begin{center}
% \small{
% \tabcolsep=0.025cm
% \begin{tabular}{|lcc|}
% \hline
% \bf Corpora &\bf Other model here & \bf Our system \\
% \hline
% \hline
% \newcite{BroTofTab2009a} corpus &1&1\\
% Buscape corpus \newcite{PangBoandLee2004} &1&1\\
% \hline
% \end{tabular}}
% \end{center}
% \caption{\changed{Accuracy (\%) on different corpora for Spanish and Portuguese systems.}}
% \label{table-supervised-models-es-pt}
% \end{table}

% Finally, \Table{} details the results 

% \subsection{Current limitations}
% 
% We showed that our system works well on long reviews. But we observed some limitations when dealing with micro-texts, such as tweets, that have become very popular in the last five years. To overcome this issue, we plan to integrate tweets parser and new compositional operation could be defined in the {\sc xml} to hanlde these texts.

% \begin{table}[t]
% \begin{center}
% \small{
% \tabcolsep=0.025cm
% \begin{tabular}{|lcc|}
% \hline
% \bf Rules &\bf SentiStrength & \bf Our system \\
% \hline
% \hline
% Baseline &1&69.85\\
% +negation &71.55\\
% +intensification &&72.35\\
% +irrealis&&\\
% \hline
% \end{tabular}}
% \end{center}
% \caption{\changed{Accuracy (\%) on different corpora for Spanish and Portuguese systems.}}
% \label{table-supervised-models-es-pt}
% \end{table}

\section{Conclusions and future work}\label{section-conclusions}

In this article, we have described, implemented and evaluated a novel model for universal and unsupervised sentiment analysis driven by a set of syntactic rules for semantic composition.
Existing unsupervised approaches are purely lexical, their rules are heavily dependent on the language concerned or
they do not consider any kind of natural language processing step in order to be able to handle different languages, using shallow rules instead.

To overcome these limitations, we introduce from a theoretical and practical point of view the concept of compositional operations, to define arbitrarily complex semantic relations between different nodes of a dependency tree.
Universal part-of-speech tagging and dependency parsing guidelines make it feasible to create multilingual sentiment analysis compositional operations that effectively address semantic composition over natural language sentences. The system is not restricted to any corpus or language, and by simply adapting or defining new operations it can be adapted to any other PoS tag or dependency annotation criteria.

We have compared our universal unsupervised model with state-of-the-art unsupervised and supervised approaches. Experimental results show: (1) that our algorithm outperforms two of the most commonly used unsupervised systems, (2) the universality of the model's compositional operations across different languages and (3) the usefulness of our approach on domain-transfer applications, especially with respect to supervised models.

As future work, we plan to design algorithms for the automatic extraction of compositional operations that capture the semantic relations between tree nodes. \changed{We would also like to collect corpora to extend our evaluation to more languages, since collections that are directly available on the web are scarcer than expected. We plan to pay special attention to Sino-Tibetan and Afro-Asiatic languages.} \changed{With respect to ungrammatical texts, we plan to integrate Tweebo parser \cite{KonSchSwaBhaDyeSmi2014a} into our system. Although it does not follow universal guidelines, it will allow us to define compositional operations specifically intended for English tweets and their particular structure}. Additionally, the concept of compositional operations is not limited to generic {\sc sa} and could be adapted for other tasks such as universal aspect extraction. Finally, we plan to adapt the \newcite{PorCamWinHua2014a} sentic patterns as compositional operations, so they can be handled universally.

\section*{Acknowledgments}

This research is supported by the  Ministerio de Econom\'{\i}a y Competitividad (FFI2014-51978-C2).
David Vilares is funded by the Ministerio de Educaci\'{o}n, Cultura y Deporte (FPU13/01180). Carlos G\'{o}mez-Rodr\'{\i}guez is funded by
an Oportunius program grant (Xunta de Galicia). We thank Roman Klinger for his help in obtaining the German data.





%\section*{References}

% \bibliographystyle{acl2016}
% \bibliography{kbs2016-universal-sa,\dropboxfolder/bib/om,\dropboxfolder/bib/misc,\dropboxfolder/bib/nlp,\dropboxfolder/bib/wordnet}

% \end{thebibliography}
\end{document}